\pdfoutput=1

\documentclass[11pt]{article}

\usepackage[final]{acl}

\usepackage{times}
\usepackage{latexsym}

\usepackage[T1]{fontenc}

\usepackage[utf8]{inputenc}

\usepackage{microtype}

\usepackage{inconsolata}

\usepackage{amssymb}
\usepackage{amsmath}
\usepackage{booktabs}
\usepackage{graphicx}
\usepackage{makecell}
\usepackage{multirow, ctable}
\usepackage{bm}
\usepackage{multirow}
\usepackage{comment}
\usepackage{nicefrac}
\usepackage{amsfonts}
\usepackage{amsthm}

\usepackage{enumitem}
\setlist[itemize,enumerate]{leftmargin=*}
\usepackage{xcolor}

\usepackage{arydshln}

\makeatletter
\def\adl@drawiv#1#2#3{%
        \hskip.5\tabcolsep
        \xleaders#3{#2.5\@tempdimb #1{1}#2.5\@tempdimb}%
                #2\z@ plus1fil minus1fil\relax
        \hskip.5\tabcolsep}
\newcommand{\cdashlinelr}[1]{%
  \noalign{\vskip 2pt
           \global\let\@dashdrawstore\adl@draw
           \global\let\adl@draw\adl@drawiv}
  \cdashline{#1}[.4pt/2pt]
  \noalign{\global\let\adl@draw\@dashdrawstore
           \vskip 2pt}}
\makeatother

\newcommand{\similarity}[1]{\textsf{sim}\left(#1\right)}

\newcommand{\bigO}[1]{\mathcal{O}\left(#1\right)}

\title{How Effective are State Space Models for Machine Translation?}

\newcommand*\samethanks[1][\value{footnote}]{\footnotemark[#1]}

\author{
 \textbf{Hugo Pitorro\thanks{~~Equal contribution.}\textsuperscript{1}},
 \textbf{Pavlo Vasylenko\samethanks\textsuperscript{2,3}},
 \textbf{Marcos Treviso\textsuperscript{3}},
 \textbf{André F. T. Martins\textsuperscript{2,3,4,5}}
\\
 \textsuperscript{1}TU Munich,
 \textsuperscript{2}Instituto Superior Técnico, University of Lisbon
 \\
 \textsuperscript{3}Instituto de Telecomunicações,
 \textsuperscript{4}Unbabel,
 \textsuperscript{5}ELLIS Unit Lisbon
}

\begin{document}
\maketitle
\begin{abstract}
Transformers are the current architecture of choice for NLP, but their attention layers do not scale well to long contexts. 
Recent works propose to replace attention with linear recurrent layers---this is the case for state space models, which enjoy efficient training and inference.
However, it remains unclear whether these models are competitive with transformers in machine translation (MT).
In this paper, we provide a rigorous and comprehensive experimental comparison between transformers and linear recurrent models for MT. 
Concretely, we experiment with RetNet, Mamba, and hybrid versions of Mamba which incorporate attention mechanisms.
Our findings demonstrate that Mamba is highly competitive with transformers on sentence and paragraph-level datasets, where in the latter both models benefit from shifting the training distribution towards longer sequences.
Further analysis show that integrating attention into Mamba improves translation quality, robustness to sequence length extrapolation, and the ability to recall named entities.  
\end{abstract}
\section{Introduction}

The inherent design of attention---the underlying mechanism of transformers---leads to quadratic computational costs and challenges in length generalization~\citep{Varis_2021}. 
As an alternative, recent works propose to replace attention with linear recurrent approaches, which enjoy efficient training and inference, and obtain competitive results in language modeling tasks~\citep{katharopoulos2020transformers, gu2022efficiently, peng-etal-2023-rwkv, sun2023retentive, gu2023mamba}.

In machine translation (MT), there is an increasing demand for supporting longer context lengths, such as paragraphs or entire documents~\citep{fernandes-etal-2021-measuring,wang-etal-2023-document-level,kocmi-etal-2023-findings}.
Given this trend, it has become increasingly important to design models capable of efficiently handling longer sequences. 
Previous research indicates that models like state space models (SSMs), exemplified by S4 \citep{gu2022efficiently}, still lag behind transformers in MT \citep{vardasbi-etal-2023-state}. 
However, it remains unclear whether these findings hold true for recent, more expressive variations of linear recurrent models, such as RetNet~\citep{sun2023retentive} and Mamba~\citep{gu2023mamba}, especially on settings that involve the use of pretrained models and long context datasets.

In this paper, we provide a rigorous and comprehensive experimental comparison between transformers, RetNet, Mamba, as well as hybrid versions of Mamba that incorporate attention mechanisms (\S\ref{sec:sentence-translation}). 
We also compare with pretrained Mamba and Pythia~\citep{biderman2023pythia} at two parameter scales, $\sim$400M and 1.4B. 
Building on existing literature that explores the capabilities of linear recurrent models in language modeling~\citep{arora2023zoology, jelassi2024repeat},
we further investigate the performance of models trained from scratch in recalling context tokens during the translation process (\S\ref{ne-recall}). 
Moreover, we extend our analysis by investigating the models' ability to handle long contexts, on paragraph-level datasets (\S\ref{sec:paragraph-translation}), along with
measuring their sensitivity to different sequence lengths (\S\ref{extrapolation-issues}) and inference cost (\S\ref{sec:inference_cost}). 
Overall, our main findings are:\footnote{\url{https://github.com/deep-spin/ssm-mt}} 

\begin{itemize}

    \item For sentence-level experiments, we show that Mamba exhibits competitive performance compared to transformers, for both trained-from-scratch and pretrained models.

    \item At the paragraph level, we find that Mamba is sensitive to the training distribution's sequence length and struggles with longer inputs. However, shifting the distribution towards longer sequence lengths helps to close the gap with transformers.

    \item We observe that integrating attention and state space models creates a strong model in terms of translation quality, robustness to sequence length extrapolation, and ability to recall named entities.

\end{itemize}

\section{Background}
\label{sec:background}

In this section, we present an overview of transformers, and the foundation  of the linear recurrent models covered in this paper: linear attention (RetNet) and state space models (Mamba).

\subsection{Transformers} 
\label{subsec:transformers}

The key component in the transformer architecture is the attention mechanism, which is responsible for contextualizing information within and across input sequences. 
Concretely, given query
$\bm{Q} \in \mathbb{R}^{n \times d}$,  key $\bm{K} \in \mathbb{R}^{n \times d}$, and  value $\bm{V} \in \mathbb{R}^{n \times d}$ matrices as input, where $n$ is the sequence length and $d$ the hidden size,
the single head \textit{self-attention mechanism} is defined as follows~\citep{vaswani2017attention}:
\begin{equation}\label{eq:attention}
    \bm{Y} = 
    \textsf{softmax}
    \Bigg(
            \frac{\bm{Q}\bm{K}^\top}{\sqrt{d}}
    \Bigg)
    \bm{V} \in \mathbb{R}^{n \times d}.
\end{equation}
For decoder-only models, a causal mask is used to ignore future tokens.
Notably, the $\bm{Q}\bm{K}^\top$ operation leads to a $\bigO{n^2}$ cost during training, and $\bigO{n}$ during inference with caching and causal masking.

\subsection{Linear Attention}
\label{subsec:linear_attention}

Denote by $\bm{q}_i, \bm{k}_i, \bm{v}_i, \bm{y}_i \in \mathbb{R}^d$ respectively the (column) vectors corresponding to the $i\textsuperscript{th}$ rows of the matrices $\bm{Q}, \bm{K}, \bm{V}, \bm{Y}$ defined above.  
\citet{katharopoulos2020transformers} reformulate the attention mechanism by casting the role of the softmax as a similarity function $\similarity {\bm{q}, \bm{k}} = \exp \left( \nicefrac{\bm{q}^\top \bm{k}}{\sqrt{d}} \right)$: %
\begin{equation}
\label{eq:linear-attn}
    \bm{y}_i = \frac{\sum_{j=1}^n \textsf{sim}(\bm{q}_i, \bm{k}_j) \bm{v}_j}
                {\sum_{j=1}^n \textsf{sim}(\bm{q}_i, \bm{k}_j)}. 
\end{equation}
However, any kernel $k(\bm{x},\bm{y}): \mathbb{R}^d \times \mathbb{R}^d \to \mathbb{R}$
is a suitable candidate for the similarity function~\citep{smola1998learning,tsai-etal-2019-transformer}. 
In particular, a kernel  $k(\bm{x},\bm{y}) = \bm{\phi}(\bm{x})^\top\bm{\phi}(\bm{y})$, where $\bm{\phi}: \mathbb{R}^d \rightarrow \mathbb{R}^r$ is a feature map, leads to:
\begin{align}
\label{eq:linear-kernel}
    \bm{y}_i &= \frac{\sum_{j=1}^{n} \bm{\phi}(\bm{q}_i)^\top \bm{\phi}(\bm{k}_j) \bm{v}_j}{\sum_{j=1}^{n} \bm{\phi}(\bm{q}_i)^\top \bm{\phi}(\bm{k}_j)} \nonumber\\
            &= \frac{\sum_{j=1}^{n} \bm{v}_j \bm{\phi}(\bm{k}_j)^\top \bm{\phi}(\bm{q}_i)}
                    {\sum_{j=1}^{n} \bm{\phi}(\bm{k}_j)^\top \bm{\phi}(\bm{q}_i)}  \nonumber\\      
            &= \frac{\bm{S}^\top \bm{\phi}(\bm{q}_i)}
                    {\bm{z}^\top \bm{\phi}(\bm{q}_i)},
\end{align}
where $\bm{S} = \sum_{j=1}^{n} \bm{\phi}(\bm{k}_j) \bm{v}_j^\top \in \mathbb{R}^{r \times d}$ and $\bm{z} = \sum_{j=1}^{n} \bm{\phi}(\bm{k}_j) \in \mathbb{R}^{r}$.
Notably, if initial states are initialized as $\bm{S}_0 = \bm{0}_{r \times d}$ and $\bm{z}_0 = \bm{0}_{r}$, intermediate states can be computed in a recurrent fashion:
\begin{align}
    \label{eq:recurrent-form}
    \bm{S}_i &= \bm{S}_{i-1} + \bm{\phi}(\bm{k}_i) \bm{v}_i^\top, \nonumber\\
    \bm{z}_i &= \bm{z}_{i-1} + \bm{\phi}(\bm{k}_i).
\end{align}
Since we can reuse the same $\bm{S}_i$ and $\bm{z}_i$ for all queries, this recurrent variant offers a $\bigO{n}$ complexity during training and enjoys a $\bigO{1}$ complexity for inference.%
\footnote{In practice, however, this recurrent view is not parallelizable, leading to chunkwise-recurrent variations for training~\citep{hua2022transformer,sun2023retentive,yang2024gated}.}

\paragraph{Retentive Networks (RetNet).} \citet{sun2023retentive} set $\bm{\phi}$ as the identity function, i.e., $k(\bm{q}, \bm{k}) = \bm{q}^\top\bm{k}$, ignore the normalizer in Equation~\ref{eq:linear-attn}, and introduce an exponential decay mask $\gamma$, leading to:
\begin{align}
    \label{eq:retnet}
    \bm{S}_i &= \gamma \bm{S}_{i-1} + \bm{k}_i \bm{v}_i^\top, \nonumber\\
    \bm{y}_i &= \bm{S}_i^\top \bm{q}_i.
\end{align}
This formulation effectively biases the attention mechanism to focus on closer token interactions. 
RetNet also uses XPos \citep{sun-etal-2023-length}, a relative positional encoding method, to improve its context extrapolation abilities.

\subsection{State Space Models (SSMs)}
SSMs \citep{gu2020hippo} provide an alternative sequence mixing layer 
by processing sequences $\bm{x}_1, ..., \bm{x}_n$, where each $\bm{x}_i \in \mathbb{R}^{d}$, through a linear recurrence.
Letting $\bm{H}_i \in \mathbb{R}^{r \times d}$ denote the ``state'' at the $i\textsuperscript{th}$ time step, a discrete SSM is defined as follows:\footnote{A discretization step is needed in order to obtain discrete parameters. For example, a possible method for this step is the zero-order hold rule, used by Mamba~\citep{gu2023mamba}.}
\begin{align}
  \label{eq:discrete-ssm}
    \bm{H}_i &= \bm{A}\bm{H}_{i-1} + \bm{b} \bm{x}_i^\top, \nonumber\\
    \bm{y}_i &= \bm{H}_i^\top \bm{c},
\end{align}
where $\bm{A} \in \mathbb{R}^{r \times r}$, $\bm{b} \in \mathbb{R}^{r}$, and $\bm{c} \in \mathbb{R}^{r}$ are (discrete) parameters.%
\footnote{The SSM equations are commonly written independenty for each input dimension $j \in [d]$ as 
\begin{align*}
\bm{h}_i^{(j)} &= \bm{A}\bm{h}_{i-1}^{(j)} + \bm{b} x_{i}^{(j)}, \quad
y_i^{(j)} = \bm{c}^\top \bm{h}_i^{(j)},
\end{align*}
with $\bm{A}$, $\bm{b}$, and $\bm{c}$ shared across input dimensions. This is equivalent to \eqref{eq:discrete-ssm}, where the $j\textsuperscript{th}$-column of $\bm{H}_i$ equals $\bm{h}_i^{(j)}$.} %
Since the same parameters are used for both relevant and irrelevant inputs, this model is deemed \emph{input-independent}, 
which, in turn, makes the model unable to reset or overwrite its hidden states. S4~\citep{gu2022efficiently} is an instance of this model, which enjoys a $\bigO{n\log n}$ time complexity during training, and $\bigO{1}$ during inference. 
\citet{vardasbi-etal-2023-state} shows that S4 still underperforms transformers for MT. 
Finally, note the similarity between Eq.~\ref{eq:retnet} and Eq.~\ref{eq:discrete-ssm}: RetNets can be seen as state space models with $\bm{A} = \gamma \bm{I}$ and data-dependent $\bm{b}$ and $\bm{c}$.

\paragraph{Mamba.} To make the SSM parameters \emph{data-dependent}, Mamba~\citep{gu2023mamba} introduces a selection mechanism that uses learnable linear projections over $\bm{x}$ prior to the discretization step, effectively making all parameters dependent on the $i\textsuperscript{th}$ input. This leads to:
\begin{align}
  \label{eq:discrete-mamba}
    \bm{H}_i &= \bm{A}_i \odot \bm{H}_{i-1} + \bm{B}_i \odot \bm{X}_i, \nonumber\\
    \bm{y}_i &= \bm{H}_i^\top \bm{c}_i,
\end{align}
where $\bm{X}_i = \bm{1}_r\bm{x}_i^\top \in \mathbb{R}^{r \times d}$ is an $r$-sized stack of the input, $\bm{A}_i \in \mathbb{R}^{r \times d}$ represents $d$ diagonal matrices of size $r \times r$, $\bm{B}_i \in \mathbb{R}^{r \times d}$, $\bm{c}_i \in \mathbb{R}^{r}$, and $\odot$ is the Hadamard product. 
Note that, unlike S4, where the same $\bm{A}$ and $\bm{B}$ parameters are shared across all hidden dimensions $1 \leq h \leq d$, 
Mamba defines $\bm{A}_i$ and $\bm{B}_i$ with a shape of $(\ldots, d)$, allowing for unique parameters in each hidden dimension.
While this formulation makes Mamba more expressive, it disrupts the convolutional approach used for training in S4. To address this, \citet{gu2023mamba} propose an efficient IO-aware and parallelizable associative scan algorithm for training~\citep{smith2023simplified}. 
Nonetheless, the recurrent view can still be used for inference with a $\bigO{1}$ time complexity.

\section{Experimental Setup}

We conduct experiments with transformers, RetNet, and Mamba for MT in \S\ref{sec:sentence-translation} and \S\ref{sec:paragraph-translation}. 
In this section, we detail the sentence and paragraph-level datasets used in our experiments, along with the settings for our models, which are trained in two distinct regimes: from scratch, or finetuned from a pretrained checkpoint.\looseness=-1
   
\subsection{Datasets}\label{sec:datasets}

For sentence-level experiments, we focus on WMT14 \textsc{de$\to$en} and WMT16 \textsc{ro$\to$en} for consistency with previous works \citep{vardasbi-etal-2023-state}, using the standard training, validation and test splits.  
For paragraph level, we use the more recent WMT23 dataset~\citep{kocmi-etal-2023-findings}, which contains $\sim$300M training samples and $\sim$1K test samples incorporating multi-sentence passages.
In order to obtain a small high-quality subset for training, we exclude ParaCrawl and CommonCrawl samples from the original dataset and clean the remaining data. 
Our cleaning process includes three steps. First, we identify and remove samples in incorrect languages via \texttt{langdetect}\footnote{\url{https://github.com/Mimino666/langdetect}}. Second, we eliminate duplicates. Third, we rank the samples using \textsc{CometKiwi-22}~\citep{rei-etal-2022-cometkiwi} a state-of-the-art translation quality estimator, and keep only the top 6M samples. We call the refined dataset WMT23-6M. 
Datasets statistics are shown in Table~\ref{tab:dataset-statistics}. \looseness=-1

\begin{table}[t]
\small
\centering
\setlength{\tabcolsep}{4pt}
\begin{tabular}{lll}
    \toprule
        \sc Dataset & \# \sc Samples & \sc \# Tokens \\  
    \midrule
        \small{IWSLT17 (\textsc{de$\leftrightarrow$en})}            & 200K & 45.2 \textcolor{gray}{$\pm$ \scriptsize{29.5}} \\
        \small{WMT16 (\textsc{ro$\leftrightarrow$en})}              & 610K & 58.9 \textcolor{gray}{$\pm$ \scriptsize{31.1}} \\
        \small{WMT14 (\textsc{de$\leftrightarrow$en})}              & 4.5M & 62.1 \textcolor{gray}{$\pm$ \scriptsize{45.6}} \\
        \small{WMT23-6M (\textsc{de$\leftrightarrow$en})}           & 6M & 58.4 \textcolor{gray}{$\pm$ \scriptsize{32.9}} \\
    \cdashlinelr{1-3}
        \small{WMT23-CAT-5 (\textsc{de$\leftrightarrow$en})}        & 2M & 171.3 \textcolor{gray}{$\pm$ \scriptsize{134.9}} \\
        \small{WMT23-CAT-10 (\textsc{de$\leftrightarrow$en})}       & 1M & 312.4 \textcolor{gray}{$\pm$ \scriptsize{282.3}} \\
    \midrule
        \small{Ted Talks Val. (\textsc{de$\leftrightarrow$en})}     & 995 & 268.5 \textcolor{gray}{$\pm$ \scriptsize{189.6}} \\
        \small{WMT23 Test (\textsc{de$\to$en})}         & 549 & 135.1 \textcolor{gray}{$\pm$ \scriptsize{147.7}} \\
        \small{WMT23 Test (\textsc{en$\to$de})}         & 557 & 185.2 \textcolor{gray}{$\pm$ \scriptsize{188.2}} \\
    \bottomrule
\end{tabular}
\caption{Sentence and paragraph-level datasets statistics.}
\label{tab:dataset-statistics}
\end{table}

\subsection{Models}

We make a broad selection of models spanning both trained-from-scratch and finetuned versions. In the first setting, we compare standard transformers, linear recurrent models, and also hybrid approaches that integrate attention into Mamba. For finetuned models, we experiment with released Pythia and Mamba checkpoints. We describe each model next. 

\subsubsection{Standard Models}

\paragraph{Transformers.} We select two variants of the transformer model as baselines: a base encoder-decoder formulation and a modern decoder-only version. The \textbf{Transformer Enc-Dec.} model, as described in the original paper \citep{vaswani2017attention}, has 77M parameters, and uses sinusoidal positional embeddings and standard ReLU activations. The second variant, \textbf{Transformer++}, is a decoder-only formulation incorporating recent advancements, such as rotary positional embeddings \citep{su2023roformer} and the SwiGLU layer \citep{shazeer2020glu}.
Specifically, we use the LLaMA architecture \citep{touvron2023llama}, adjusting the embedding dimension to match the parameter count of the base transformer (79M), consistent with the version employed in \citep{gu2023mamba}.

\paragraph{Linear recurrent models.} We select two representative recurrent models, \textbf{RetNet} \citep{sun2023retentive} and \textbf{Mamba} \citep{gu2023mamba}.
Both models are tested with 77M parameters to approximately match the number of parameters in the transformer models.

\subsubsection{Hybrid Models}

Previous work has shown that incorporating attention into linear recurrent models leads to strong performance in language modeling \citep{ fu2023hungry, arora2024simple, de2024griffin}. Therefore, we aim to examine if this is also the case for MT by exploring three hybrid variants, detailed next.

\paragraph{Mamba-MHA.} The simplest hybrid formulation involves replacing some of the Mamba layers with attention. Some natural questions then arise: how many attention layers are needed, and where to place them? 
After careful ablations, detailed in Appendix~\ref{appendix:hybrid-ablations}, we use two attention layers placed at the middle and at the output of the network, resembling the hybrid version of H3~\citep{fu2023hungry}.

\paragraph{Mamba-Local.}  While aiming to achieve robust performance, the introduction of full attention to Mamba disrupts its efficiency gains. Thus, we consider local attention variants such as sliding window attention \citep{beltagy2020longformer,child2019generating}, employed in recent hybrid models \citep{arora2024simple, de2024griffin}.
We use a window size of $64$ based on the average sequence length shown in Table~\ref{tab:dataset-statistics} and ablations in Appendix~\ref{appendix:hybrid-ablations}. 

\paragraph{Mamba Enc-Dec.} Lastly, inspired by the S4-based encoder-decoder model from \citet{vardasbi-etal-2023-state}, we replace the self-attention mechanism in transformers with a Mamba block and keep the cross-attention component intact. 
In terms of complexity, since this variant computes attention over the source sentence, it incurs an additional $\bigO{n^2}$ cost for training and $\bigO{n}$ for inference.

\subsubsection{Pretrained Models}

In order to fairly evaluate the relative performance between pretrained models, we need to ensure consistency between their pretraining data. 
Taking this into account, we consider two strong models pretrained on The Pile \citep{gao2020pile}: Pythia~\citep{biderman2023pythia}, a modern transformer, and Mamba, a modern SSM.
Note, however, that Pythia was pretrained on more tokens than Mamba (see Table~\ref{tab:pretrained-details}), hence the comparison might be slightly unfavorable to Mamba.
We experiment with two model scales, \emph{small} (S) and \emph{medium} (M).
Concretely, we experiment with Pythia 410M and 1.4B, and with Mamba 370M and 1.4B.

\begin{table*}[t]
\centering

\setlength{\tabcolsep}{4pt}
\begin{tabular}{ll c@{\,\;\;}cc c@{\;}cc c@{\;}cc c@{\;}cc}
    \toprule
        & & & 
        \multicolumn{5}{c}{WMT16} & & \multicolumn{5}{c}{WMT14} \\
        \cmidrule(lr){4-8} \cmidrule(lr){10-14}
        
        & & & 
        \multicolumn{2}{c}{\sc ro$\to$en} & & 
        \multicolumn{2}{c}{\sc en$\to$ro} & & 
        \multicolumn{2}{c}{\sc de$\to$en} & & 
        \multicolumn{2}{c}{\sc en$\to$de} \\
        \cmidrule(lr){4-5} \cmidrule(lr){7-8} \cmidrule(lr){10-11} \cmidrule(lr){13-14}
        \sc Model & \sc Size & & \sc bleu & \sc comet & & \sc bleu & \sc comet & & \sc bleu & \sc comet & & \sc bleu & \sc comet \\

    \midrule    
        \multicolumn{10}{l}{\textit{Trained from scratch}} \\
        Transf. Enc-Dec     & 77M & &\textbf{29.2} & \underline{74.8} & & \underline{22.0} & \textbf{78.6} & & 27.4 & 78.6 & & 22.3 & 77.1 \\
        Transformer++       & 79M & & 26.4 & 72.6 & & 21.7 & 72.7 & & 26.9 & 79.0 & & 22.8 & 77.9 \\
        RetNet              & 77M & & 26.4 & 72.4 & & 19.9 & 76.0 & & 23.4 & 74.7 & & 19.6 & 71.7 \\
        Mamba               & 77M & & 27.0 & 73.8 & & 21.4 & 77.9 & & \textbf{27.5} & 80.2 & & 22.4 & 77.8 \\
    \cdashlinelr{1-14}
        Mamba-MHA           & 78M & & \underline{28.5} & \textbf{75.1} & & 21.7 & 78.3 & & \underline{27.4} & \textbf{80.6} & & \textbf{23.2} & \textbf{79.9} \\
        Mamba-Local         & 78M & & 25.9 & 73.9 & & 20.9 & 76.9 & & 27.2 & \underline{80.1} & & \underline{23.2} & \underline{79.5} \\
        Mamba Enc-Dec       & 82M & & 28.5 & 74.4 & & \textbf{22.7} & 77.9 & & 27.2 & 80.0 & & 21.6 & 78.8 \\
    \midrule
        \multicolumn{10}{l}{\textit{Finetuned}} \\

        Pythia-S        & 410M  & & 33.4 & 82.0 & & 24.1 & 85.8 & & 30.9 & 83.6 & & 25.2 & 84.0 
        \\
        Mamba-S         & 370M  & & \textbf{34.1} & \textbf{83.2} & & 24.2 & \underline{86.4} & & 29.8 & 83.3 & & 25.0 & 83.2 \\
        
        Pythia-M        & 1.4B  & & \underline{33.9} & \textbf{83.2} & & \textbf{24.9} & \textbf{87.1} & & \textbf{32.2} & \textbf{84.5} & & \textbf{26.7} & \textbf{84.9}
        \\
        Mamba-M         & 1.4B  & & 33.8 & 83.1 & & \underline{24.5} & 86.2 & & \underline{31.9} & \textbf{84.5} & & \underline{26.5} &  \underline{84.2}\\
    
    \bottomrule
    
\end{tabular}
\caption{Sentence-level results in terms of BLEU and COMET for models trained from scratch (top) and models finetuned from a pretrained checkpoint (bottom). \textbf{Bold} represents top results; \underline{underline} represents second-best.}
\label{tab:sentence_performance_scores}
\end{table*}

\subsection{Training and Evaluation}

For models trained from scratch, we follow the settings proposed in \citep{vardasbi-etal-2023-state}, whereas for pretrained models, we follow the finetuning settings used by Mamba~\citep{gu2023mamba}.
For decoder-only models, we pass a concatenation of the source and target sequences separated by a special token as input.
We evaluate all models with BLEU \citep{post-2018-call}\footnote{SacreBLEU signature: \texttt{|1|mixed|no|13a|exp|}} and COMET \citep{rei-etal-2022-comet}.\footnote{\href{https://huggingface.co/Unbabel/wmt22-comet-da}{\texttt{huggingface.co/Unbabel/wmt22-comet-da}}}
We base our analysis on the latter, given its strong correlation with human judgments on sentence and paragraph-level data~\citep{freitag-etal-2022-results,freitag-etal-2023-results}. 
More training details can be found in \S\ref{appendix:training-details}.\looseness=-1

\section{Sentence-level Translation}\label{sec:sentence-translation}

We start by evaluating our standard, hybrid, and finetuned models on the sentence-level WMT16 \textsc{ro$\leftrightarrow$en} and WMT14 \textsc{de$\leftrightarrow$en} datasets.
Results can be found in Table~\ref{tab:sentence_performance_scores} in terms of BLEU and COMET. Next, we discuss the key findings.

\subsection{Discussion} \label{sec:discussion_sentence_translation}

\paragraph{Mamba is competitive when trained from scratch.} 
Mamba, a decoder-only model, not only outperforms a decoder-only transformer (Transformer++) across the board, but also an encoder-decoder transformer (Transf. Enc-Dec) in the larger WMT14 for both \textsc{de$\leftrightarrow$en} language pairs. 
This creates a contrast with the S4 results obtained by \citet{vardasbi-etal-2023-state}. 
We hypothesize that Mamba's good results are due to its data-dependent state updates (Eq.~\ref{eq:discrete-mamba}), which allows for more precise information retention in its hidden state. 
On the other hand, RetNet's performance is generally subpar compared to other models, likely due to its strong locality bias (induced by $\gamma$ in Eq.~\ref{eq:retnet}), which may hinder performance in MT, 
a task where the source input servers as a prefix to the translation, and it requires ``focused attention'' during decoding.

\paragraph{Attention benefits Mamba.} 
By including attention layers in Mamba's architecture, we find that Mamba-MHA, which employs only two attention layers, is able to outperform both transformers and Mamba for almost all language pairs. 
While Mamba-Local retains constant inference complexity via windowed attention, it is not as strong as the full attention variant. 
Finally, Mamba Enc-Dec's performance is also competitive,  falling just short of Mamba-MHA and echoing the S4 encoder-decoder findings of \citet{vardasbi-etal-2023-state}. 

\paragraph{Pretraining improves all models.}
We note a large COMET gap, roughly 4-8 COMET points, between the finetuned models and those trained from scratch for all language pairs. 
This is expected, since not only are these models bigger, but they also have strong data-driven priors, which are beneficial in downstream tasks~\citep{amos2024never}. 

\paragraph{Larger models achieve top results.}
For small models, Mamba outperforms Pythia for \textsc{ro$\leftrightarrow$en} in terms of COMET and BLEU. However, Pythia is superior on the larger \textsc{de$\leftrightarrow$en} datasets.
Moving to larger models, we note that Mamba improves COMET scores by $\sim$1 point on \textsc{en$\leftrightarrow$de} while dropping only 0.1-0.2 on \textsc{en$\leftrightarrow$ro} datasets. 
On the other hand, Pythia improves results consistently for all language pairs with a larger model size, outperforming or matching the results of other models.
On average, 
we find that both their gaps decrease when moving from smaller to medium-sized models but Pythia benefits more in terms of COMET.
It is worth noting that Mamba is pretrained on fewer samples than Pythia (see Table~\ref{tab:pretrained-details}) and that the impact of pretraining data quality can also play a role in downstream task performance.

\subsection{Recall of Named Entities}\label{ne-recall}

\begin{figure*}[t]
    \centering
    \includegraphics[width=\textwidth]{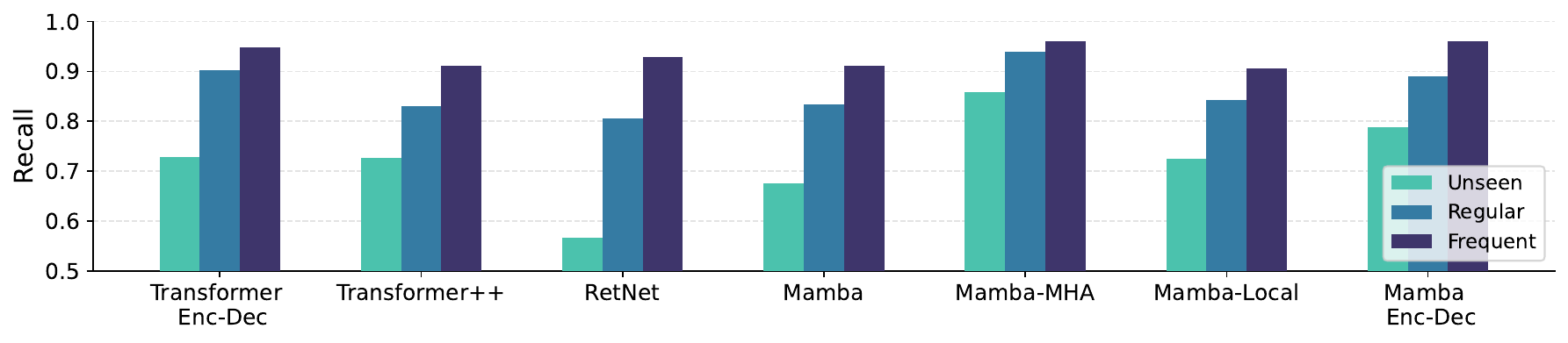}
    \captionof{figure}{Recall in recovering named entities on the WMT16 \textsc{ro$\to$en} dataset by their training set frequency: \textit{unseen} entities do not appear in the training data, while \textit{regular} and \textit{frequent} entities appear $[1, 16)$ and $16+$ times.}
    \label{fig:ne-recall-frequency}
\end{figure*}

Following our discussion of sentence-level translation, we now focus on how well these models recall context tokens during translation. 
Inspired by prior studies investigating the recall of context tokens in language modeling with state space models~\citep{arora2023zoology, jelassi2024repeat}, we conduct a similar experiment for MT. 
Unlike language modeling, where token patterns often recur within a near context, MT presents a challenge due to the distinct spelling of words across languages. Therefore, we focus on the recall of named entities (NEs) that appear verbatim in both source and target sentences, using NLTK for NE recognition~\citep{bird-2006-nltk}.

We assess the models' ability to recall NEs from the WMT16 \textsc{ro$\to$en} dataset according to their frequency in the training set, as illustrated in Figure~\ref{fig:ne-recall-frequency}. 
The results reveal a clear correlation between NE frequency and their chance to be recalled in the translation process, as more frequent NEs are recalled more often. 
Notably, we note a disparity in performance with unseen entities, which provides a more illustrative view of recall ability. 
In this respect, transformers and Mamba perform on par, while RetNet shows inferior results. 
As before, the hybrid models are promising, with Mamba-MHA outperforming all models, followed closely by Mamba Enc-Dec.

\section{Paragraph-level translation}
\label{sec:paragraph-translation}

While Mamba shows competitive performance with transformers on sentence-level datasets (see Table~\ref{tab:sentence_performance_scores}), it was originally designed to handle long sequences.
Thus, we now turn our attention to paragraph-level datasets.
This allows us to study the models' sensitivity to long sequence lengths along with their robustness in handling sequences that are longer than the ones seen during training.%
\footnote{We dropped RetNet and Mamba-Local as they already achieve poor results on long \emph{sentence-level} inputs (see Fig.~\ref{fig:wmt14-comet-lens}).} 

To this end we focus on the WMT23-6M dataset (\S\ref{sec:datasets}), from which the training and test sets are composed of sentence and paragraph-level passages, respectively. 
In order to see the impact of training on long sequences, we propose to artificially construct multi-sentence datasets, which we call WMT23-CAT-$N$. Our procedure is as follows:
\begin{enumerate}
    \item We first retain the original training samples from WMT23-6M with a probability of $50\%$. 
    \item Next, for the remaining part, we concatenate $N$ random training samples.
\end{enumerate}
This approach ensures that we consistently maintain a $50\%$ ratio between single-sentence and multi-sentence samples. 
For validation, we sample 1-to-10-sentence passages from the TED Talks dataset~\citep{cettolo-etal-2012-wit3}.  
Statistics for CAT-$N$ datasets can be found in Table~\ref{tab:dataset-statistics}.  
COMET scores on the WMT23 \textsc{en$\leftrightarrow$de} test sets are shown in Table~\ref{tab:paragraph_performance_scores}. We provide additional BLEU scores in Table~\ref{tab:paragraph-mt-results} in Appendix~\ref{appendix:full-paragraph}. 
Next, we discuss our key findings.

\begin{table*}[t]
\centering
\setlength{\tabcolsep}{4pt}
\begin{tabular}{ll l@{} ccc c@{} ccc}
    \toprule
        & & &
        \multicolumn{3}{c}{\textsc{de$\to$en}} & & \multicolumn{3}{c}{\textsc{en$\to$de}} \\
        \cmidrule(lr){3-6} \cmidrule(lr){7-10} 
        
        \sc Model & \sc Size & &
        \sc orig. & \sc cat5 & \sc cat10 & &
        \sc orig. & \sc cat5 & \sc cat10 \\

    \midrule    
        \multicolumn{8}{l}{\textit{Trained from scratch}} \\
        Transf. Enc-Dec     & 77M & & 72.4 & 74.6 & 69.6 & & 65.2 & \underline{70.3} & \underline{70.3} \\
        Transformer++       & 79M & & 70.7 & 73.6 & 72.8 & & 64.8 & 69.1 & 68.8 \\
        Mamba               & 77M & & 70.0 & 73.3 & 72.3 & & 63.3 & 67.5 & 67.8 \\
    \cdashlinelr{1-10}
        Mamba-MHA           & 78M & & 72.7 & 74.2 & \underline{74.5} & & 67.0 & 68.6 & 69.7 \\
        Mamba Enc-Dec       & 82M & & 70.7 & 73.8 & \bf 75.6 & & 65.3 & \bf 71.0 & 70.1 \\
    \midrule
        \multicolumn{8}{l}{\textit{Finetuned}} \\
        Pythia-S        & 410M & & 77.4 & 78.4 & 79.0 & & 76.7 & \underline{77.8} & 77.1 \\
        Mamba-S         & 370M & & 77.2 & 78.2 & 78.3 & & 72.4 & 74.2 & 73.1 \\
        Pythia-M        & 1.4B & & 76.2 & 78.6 & 79.4 & & 75.8 & 77.4 & \bf 79.0 \\
        Mamba-M         & 1.4B & & 74.6 & \bf 79.6 & \underline{79.5} & & 73.4 & 77.5 & 77.3 \\
    \bottomrule
\end{tabular}
\caption{Paragraph-level results in terms of COMET for models trained from scratch (top) and models finetuned from a pretrained checkpoint (bottom) on WMT23 \textsc{en$\leftrightarrow$de} test set, according to the training dataset: original WMT23-6M, WMT23-CAT-5 and WMT23-CAT-10.
\textbf{Bold} represents top results; \underline{underline} represents second-best.}
\label{tab:paragraph_performance_scores}
\end{table*}

\subsection{Discussion} \label{sec:discussion_paragraph_translation}

\paragraph{Concatenation helps.}
Our strategy of concatenating sentences proves beneficial for almost all models, as COMET scores tipically improve with the CAT-5 and CAT-10 datasets, whether models are trained from scratch or finetuned.
Among models trained from scratch, Transformer Enc-Dec, Mamba-MHA, and Mamba Enc-Dec show substantial improvements, with Mamba Enc-Dec achieving the best overall results. 
For finetuned models, concatenation benefits larger models; Mamba-M outperforms Pythia-M in \textsc{de$\to$en} but underperforms in \textsc{en$\to$de}. 
Interestingly, for both training regimes, the concatenation strategy can lead to COMET gains of up to 5 points.

\paragraph{Finetuning outperforms training from scratch.}
Finetuned models consistently achieve higher COMET scores, with larger models attaining the top results overall. 
Similar to sentence-level experiments, Pythia outperforms Mamba when trained on the original, WMT23-6M dataset. 
However, both Pythia and Mamba benefit from our concatenation strategy. 
While these results indicate that our concatenation strategy helps in translating long inputs, it remains unclear whether performance on short inputs is compromised or if the models can handle longer inputs than those seen during training. We investigate these issues next.

\subsection{Sensitivity to Input Length}\label{extrapolation-issues}

\begin{figure*}[htb!]
    \centering
    \includegraphics[width=\textwidth]{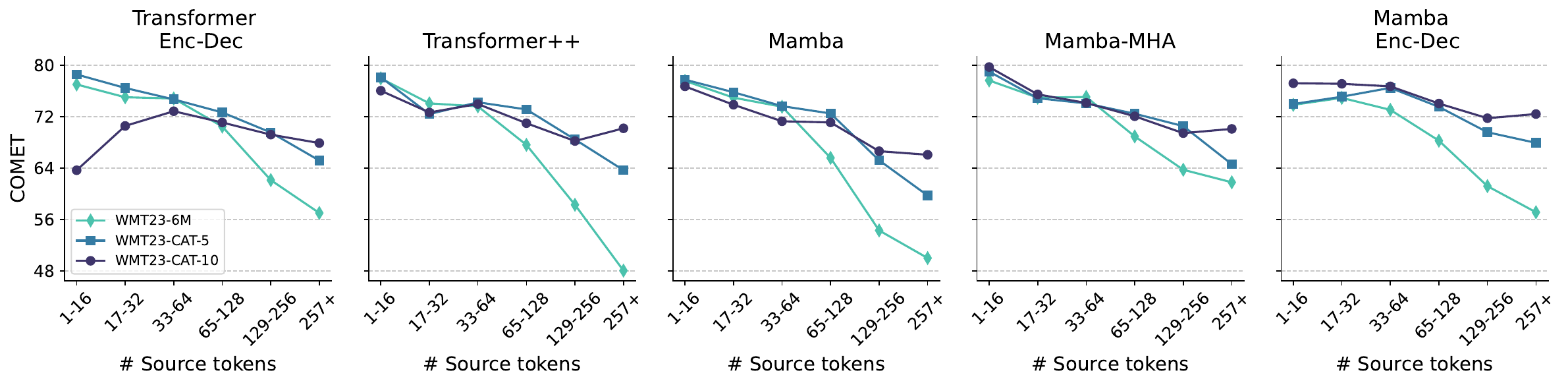}
    \includegraphics[width=0.96\textwidth]{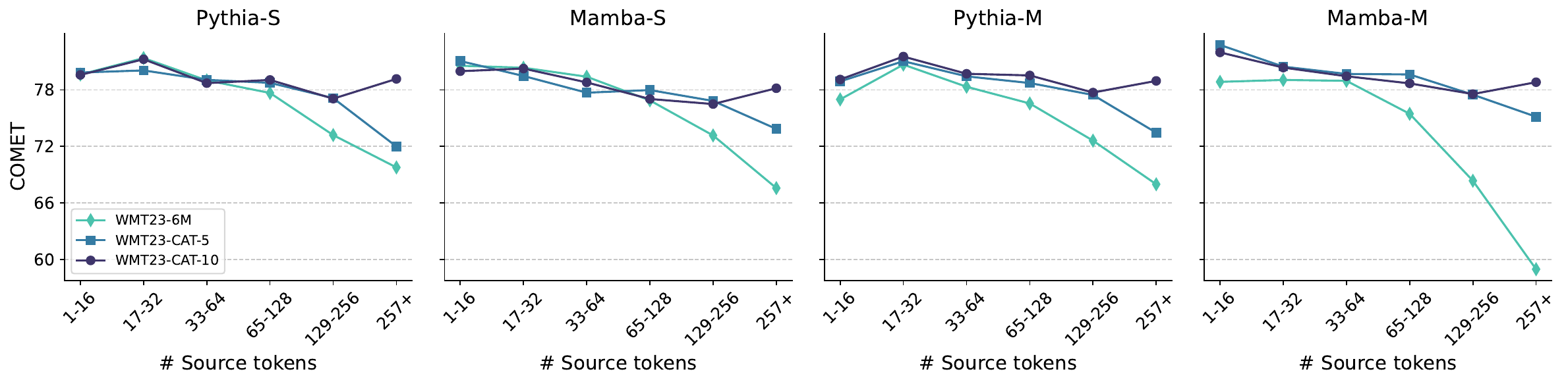}
    \captionof{figure}{Sensitivity to input length, measured by the number of sources tokens, on the WMT23 \textsc{de$\to$en} datset, for models trained from scratch (top) and finetuned from a pretrained checkpoint (bottom).
    }
    \label{fig:scratch-ms-extrapolation}
\end{figure*}

Based on the previous observations, we notice that performance between models varies considerably after being exposed to different sequence lengths during training. 
In this subsection, we investigate how robust each model is to length distribution shifts between training and test.  
Results are shown in Figure~\ref{fig:scratch-ms-extrapolation} for both training regimes on the WMT23 \textsc{de$\to$en} dataset. Results are consistent for \textsc{en$\to$de}, shown in Figure~\ref{fig:ende-ms-extrapolation}, Appendix~\ref{sec:sensitivity-seq-len}.

\paragraph{Discussion.}
When training on WMT23-6M, we observe a decline in performance for all models on long sequences, affecting both trained-from-scratch and finetuned variants.
Interestingly, this degradation is evident in Mamba, as expected due to its finite hidden state capacity. However, it is also challenging for transformers despite their relative positional embeddings.
Moreover, both finetuned and hybrid models exhibit more consistent performance across different sequence lengths, even on the original sentence-level dataset, suggesting a more robust capability for dealing with long-context inputs.

Overall, our concatenation approach has largely mitigated the performance issues with long inputs present in models trained on WMT23-6M, as models trained on CAT datasets produce higher-quality translations for longer sequences.
This improvement is uniform across all models, with CAT-10 yielding consistently better translations in the longest bin (257+ tokens). 
However, the CAT-10 dataset seems to reduce translation quality for shorter samples in some models, though this effect is minimal or absent in hybrid and finetuned models. 
We further examine the ability to extrapolate to even longer sentences (up to 2048 tokens) than those seen during training in \S\ref{sec:sensitivity-seq-len}, finding that finetuned versions of Mamba are more robust than Pythia when trained on CAT-10.

\subsection{Inference Cost}
\label{sec:inference_cost}

In \S\ref{sec:background} we covered the theoretical time complexity of our models 
in training and inference time. 
Here, we examine the wallclock time and memory usage of Pythia and Mamba in a realistic setting 
where inputs are WMT23 \textsc{de$\to$en} test samples, %
and outputs continue to be generated until they reach exactly $L \in \{512, 1024\}$ tokens. 
Table~\ref{tab:inference-time-and-memory} shows that 
Mamba's memory usage is significantly lower than Pythia's, with gaps of $\sim$ 3-5x overall.
The wallclock time difference is not as notable but still substantial, especially for larger models, where Mamba-M is $2$x faster than Pythia-M for $L=1024$. 
In other words, Mamba-M throughputs $\sim$806 tokens/s while Pythia-M outputs $\sim$405 tokens/s, aligning with \citep{gu2023mamba}.\footnote{Computed as $\text{batch-size} \times L / \text{wallclock-time}$.}

\begin{table}[t]
    \centering
    \small
    \begin{tabular}{l l@{\ } rr c@{\ } rr} 
    \toprule
    & & \multicolumn{2}{c}{512} & & \multicolumn{2}{c}{1024} \\
    \cmidrule(lr){2-4} \cmidrule(lr){5-7} 
    \sc Model & & 
    \sc T (s) & \sc M (GB) & & 
    \sc T (s) & \sc M (GB)  \\
    \midrule
   
    Pythia-S & & 11.52 & 2.472 & & 25.80 & 3.934 \\
Mamba-S & & 10.38 & 0.839 & & 20.59 & 1.607  \\
Pythia-M & & 14.88 & 4.789 & & 40.41 & 7.841  \\
Mamba-M & & 10.29 & 0.913 & & 20.31 & 1.668  \\
    \bottomrule
    \end{tabular}

    \caption{Average time (T) and maximum allocated memory (M) of 30 inference runs with batch size 16 on WMT23 \textsc{de$\to$en}.}

    \label{tab:inference-time-and-memory}
\end{table}

\section{Related Works}

\paragraph{Linear recurrent models for MT.} Our work is closely related to \citep{vardasbi-etal-2023-state}, which compares SSMs and transformers.
Furthermore, they also experiment with hybrid architectures composed of S4 and attention layers, an approach that has since become common~\citep{arora2024simple, de2024griffin, glorioso2024zamba}. 
In this work, we experiment with more recent linear recurrent models and their respective hybrid versions while also including larger and pretrained variants. 
Our analysis further includes investigating each model's ability to recall named entities, along with measuring translation performance across different sequence lengths on paragraph-level datasets.
In contrast to \citet{vardasbi-etal-2023-state}'s results showing that S4 lags behind transformer baselines in MT tasks, we observe that Mamba, a modern SSM, is competitive with transformers on sentence and paragraph-level datasets, whether trained from scratch or fine-tuned from a pretrained checkpoint, especially in the first setting when equipped with attention mechanisms.

\paragraph{Linear recurrent models' limitations.} 
Recent works show that Mamba struggles in tasks that involve recalling context tokens~\citep{arora2023zoology, jelassi2024repeat}, such as the synthetic Multi-Query Associative Recall task. 
In MT, however, context tokens (source and translation prefix) are not often replicated in the output (translation). 
In this work, we study this phenomenon with named entities and analyze the recall ability of transformers and linear recurrent models in \S\ref{ne-recall}.

\paragraph{Sentence concatenation} \citet{kondo-etal-2022-japanese,Varis_2021} analyze transformers' generalization towards sequence length. 
They show that transformers are susceptible to the training distribution of context length and that concatenating multiple sentences can improve the translation of longer sentences. 
Specifically, \citet{kondo-etal-2022-japanese} augment the original data with samples containing concatenations of two random sentences, while \citet{Varis_2021} concatenate up to six sentences. 
While these studies focused on sentence-level translation with sequence lengths up to 120 tokens, in this work, we extend the analysis to much longer sequences and test on paragraph-level data from the WMT2023 dataset.

\section{Conclusion}

We set out to evaluate recent linear recurrent models, particularly RetNet and Mamba, in MT tasks while thoroughly comparing them to transformer baselines and hybrid models, which combine Mamba and attention. 
We find that Mamba models are competitive with transformers, both when they are trained from scratch and when they are finetuned from a pretrained checkpoint; however, the performance delta is smaller in the latter regime. 
Our paragraph-level experiments reveal that models are hindered by the mismatch in the training and test length distributions; however, a simple concatenation approach helps to mitigate the issue.
We find that hybrid models are only slightly affected by this issue while also being competitive or outperforming transformers.
Finally, we note that Mamba models also exhibit a faster runtime and consume less memory than transformers.

\section*{Acknowledgments}

We thank Haau-Sing Li, Saul Santos, Patrick Fernandes, Sweta Agrawal and Nuno Guerreiro for their useful and constructive comments. 
This work was supported by the Portuguese Recovery and Resilience Plan through project C645008882-00000055 (Center for ResponsibleAI), by the EU’s Horizon Europe Research and Innovation Actions (UTTER, contract 101070631), by the project DECOLLAGE (ERC-2022-CoG 101088763), and by Fundação para a Ciência e Tecnologia through contract UIDB/50008/2020.

\section*{Limitations}

We point out some limitations of the presented study. First, one limitation is that we refrain from pretraining the hybrid models due to the high associated compute costs. To this effect, while our trained-from-scratch results are promising, validating them with a larger scale and strong language priors would strengthen our claim of their good performance. 
Secondly, our experiments (\S\ref{extrapolation}) appear to indicate larger models are more robust to sequence length issues. 
Nonetheless, we limited our study to models with parameter scales between 370M and 1.4B since, in preliminary sentence-level experiments, translation quality gains plateaued at the latter scale.

In another direction, we mainly rely on automated metrics for evaluating translation quality, which might not fully capture the accuracy of the translation. 
We alleviate this fault by considering the recollection of NEs in translations (\S\ref{ne-recall}). 
Furthermore, our experiments in \S\ref{extrapolation-issues} do not have a notion of translation difficulty, which might help explain the differences between models and associated datasets in different length buckets (albeit sentence length and difficulty may be connected).

\section*{Potential Risks}
Translation biases and error modes inherent in transformed-based LLMs could also be manifested in the linear recurrent models studied in this paper. 
Careful evaluation and mitigation strategies, such as detecting and overcoming hallucinations~\citep{guerreiro-etal-2023-looking,dale-etal-2023-detecting}, can alleviate these risks and ensure models' responsible use.
It should also be noted that although SSMs are potentially more energy efficient than transformer-based models, they still pose energy consumption concerns, particularly due to the large size of the models.

\bibliography{anthology_small,custom}

\newpage

\appendix

\section{Implementation and Training Details}\label{appendix:training-details}

All experiments were carried on Nvidia RTX A6000 GPUS with 48GB VRAM, and the training framework is constructed around PyTorch Lightning.\footnote{\url{https://lightning.ai/docs/pytorch/}} 
To train and generate translations in batches, we use a left-padding strategy. However, for Mamba, additional functionality is required to avoid processing padding tokens. 
To address this, we zero out inputs before and after convolution at the positions of the padding tokens and sacrifice some efficiency by using the slow path in Mamba\footnote{\url{https://github.com/state-spaces/mamba/issues/216}}. Notably, during inference, the slow path affects only the initial processing of the prompt and does not impact the actual generation.
Moreover, we added Dropout~\citep{srivastava2014dropout} to Mamba blocks, which was missing in the original implementation. Specifically, dropout is applied after the last linear projection of the Mamba block. 
Additionally, following the findings in \citep{vardasbi-etal-2023-state}, we calculate cross-entropy loss only for target tokens. 
During training, we use greedy decoding and select the top model using BLEU as the validation metric, as it is faster to compute in comparison to COMET.
For inference, we use beam search with a beam size of 5. Due to the time-consuming nature of our experiments, we report the results of a single run for all experiments.
The overall model structure and hyperparameters across both training regimes, from-scratch (\S\ref{sec:from-scratch}) and finetuning (\S\ref{sec:finetuning}), are shown in Table~\ref{tab:hyperparameters}.
Furthermore, all models were trained with \verb|bfloat16| precision. 

\begin{table}[t]
\small
\centering
\setlength{\tabcolsep}{4pt}
\begin{tabular}{lcccccc}
    \toprule
        \sc Model & \sc Size & LR & L & H & D & FFN \\
    \midrule
        \multicolumn{5}{l}{\textit{Trained from scratch}} \\
        Transf. Enc-Dec & 77M & 4e-4 & 6-6 & 8 & 512 & 2048  \\
        Transf.++ & 79M & 4e-4 & 12 & 8 & 496 & 1984 \\
        RetNet & 77M & 1e-3 & 12 & 4 & 512 & 1024 \\
        Mamba & 77M & 1e-3 & 24 & - & 610 & - \\
    \cdashlinelr{1-7}
        Mamba-MHA & 78M & 7e-4 & 24 & 8 & 624 & -  \\
        Mamba-Local  & 78M & 7e-4 & 24 & 8 & 624 & - \\
        Mamba Enc-Dec & 82M & 7e-4 & 8-6 & 8 & 512 & 2048 \\
    \midrule
        \multicolumn{5}{l}{\textit{Finetuned}} \\
        Pythia-S & 410M & 1e-5 & 24 & 16 & 1024 & 4096 \\
        Mamba-S & 370M & 3e-4  & 24 & - & 1024 & - \\
        Pythia-M & 1.4B & 1e-5 & 24 & 16 & 2048 & 8192 \\
        Mamba-M & 1.4B & 3e-4 & 24 & - & 2048 & - \\
    \bottomrule
\end{tabular}
\caption[Hyperparameters]{Detailing the full set of hyperparameters for the different models. Encoder-Decoder models have their number of layers separated by each module. LR represents the Learning Rate; L represents the number of layers; H is the number of Attention Heads; D is the model dimension; FFN is the size of the inner feed-forward network. 
}
\label{tab:hyperparameters}
\end{table}

\subsection{Training from Scratch}\label{sec:from-scratch}

Regarding tokenization, we leverage the HuggingFace \textit{tokenizers} library\footnote{\url{https://github.com/huggingface/tokenizers} \label{fnote:hf-tokenizers}} and construct a separate BPE tokenizer \citep{sennrich-etal-2016-neural} per dataset. The total vocabulary size is 32000 tokens.
We carried out a hyperparameter search to select appropriate dropout values, learning rates and architectural decisions, with the latter two detailed in Table~\ref{tab:hyperparameters}. We employ a dropout of $0.3$ and $0.1$ for both WMT14 and the different variations of WMT23. Other hyperparameters were kept intact. Concretely, we use the Inverse Square Root learning rate scheduler~\citep{vaswani2017attention} with 4000 warmup steps and a weight decay of $0.001$.

\subsection{Finetuning Pretrained Checkpoints}\label{sec:finetuning}

We employ pretrained models and corresponding tokenizers from the Huggingface library. Table \ref{tab:pretrained-details} shows the number of tokens and the size of the context window used during pretraining. For finetuning, in all experiments, we use a dropout of 0.1 with the exception of WMT16 \textsc{en$\leftrightarrow$ro}, where dropout varies from 0.1 to 0.3. Moreover, we use weight decay only in Mamba-M, with a value of $2 \cdot 10^{-4}$. Additionally, learning rates and models' attributes are shown in Table  \ref{tab:hyperparameters}.

\begin{table}[t]
    \small
    \centering
    \begin{tabular}{llll} 
    \toprule
      \sc Model   & \sc Size & \makecell{\sc Training \\ \sc tokens} & \makecell{\sc Context \\ \sc tokens} \\ 
    \midrule
         Pythia-S & 410M&  300B & 2048 \\
 Pythia-M & 1.4B& 300B & 2048 \\
         Mamba-S & 370M&  7B & 2048\\
 Mamba-M & 1.4B & 26B & 2048\\
 \bottomrule
    \end{tabular}
    \caption{Pre-training details. All models were pretrained on The Pile~\citep{gao2020pile}.}
    \label{tab:pretrained-details}
\end{table}

\subsection{Inference Cost}

For the inference cost experiments, we measure 
overall wallclock time using cuda events and  cuda synchronization from \texttt{torch.cuda} module. The overall time includes the entire generation pipeline.
Moreover, we use \texttt{torch.cuda.max\_memory\_allocated} to measure memory usage.

\section{Hybrid Models Ablation}\label{appendix:hybrid-ablations}
Building on the shortcomings of linear models \citep{akyürek2024incontext, arora2023zoology, jelassi2024repeat}, we designed hybrid models to complement SSMs with attention mechanisms.
In this section, we ablate the design choices leading to the construction of our hybrid models.
These experiments were conducted using the IWSLT17 \textsc{de$\leftrightarrow$en} dataset~\citep{cettolo-etal-2017-overview}. Results are shown in Table~\ref{tab:hybrid-ablations}.

\begin{table}[t]
\small
\centering
\setlength{\tabcolsep}{4pt}
\begin{tabular}{lcccc}
\toprule
    &  \multicolumn{2}{c}{\textsc{de$\to$en}} & \multicolumn{2}{c}{\textsc{en$\to$de}}  \\
    \cmidrule(r){2-3} \cmidrule(lr){4-5} 
         & BLEU & COMET  & BLEU & COMET \\
    \midrule
        \multicolumn{5}{l}{\textit{Mamba-MHA}} \\
        Interleaved & 30.81 & 77.98 & 24.40 & 72.48 \\ 
        L1,11 & 30.52 & 78.10 & \textbf{24.99} & 73.76\\ 
        L11,23 & \textbf{30.81} & \textbf{78.30}  & 24.40  & \textbf{73.94} \\
    \midrule
        \multicolumn{5}{l}{\textit{Mamba-Local}} \\
        Interleaved - w64 & 28.85 & 76.76 & 23.61 & 72.10 \\ 
        L11,23 - w16  & 29.37 & 77.19 & 24.12 & 72.88 \\
        L11,23 - w32  & 28.24 & 76.44 & 23.20 & 72.22 \\ 
        L11,23 - w64  & 29.40 & 77.56 & 24.41 & 72.98 \\ 
        L11,23 - w128  & 30.49 & 77.98 & 24.85 & 73.58 \\ 
\bottomrule
\end{tabular}
\caption{Hybrid models ablations with BLEU and COMET scores on the IWSLT17 dataset. Different window sizes are denoted as w$\{16,32,64,128\}$. \textit{Interleaved} refers to alternating Mamba and attention layers. L\textit{1,11} and L\textit{11,23} refer to placing attention in layers $2$ - $N/2$ and $N/2$ - $N$, respectively.}
\label{tab:hybrid-ablations}
\end{table}

Since our Mamba-MHA model replaces a set of Mamba layers with attention modules, we ablated various configurations to determine the optimal number and placement of attention layers.
Our analysis of COMET scores indicated that incorporating two attention layers significantly boosted performance, aligning with findings in previous studies \citep{fu2023hungry}. The placement of these layers had a minimal effect, leading us to select the configuration with layers at positions  $N/2$ and $N$ for further experiments due to its consistently higher COMET scores.

In the case of Mamba-Local, which uses a sliding window attention, we explored various window sizes.
Our experiments revealed that performance generally improved with window size in a linear way. 
Ultimately, a 128-token window nearly matched full attention performance, and two layers of 64-token windowed attention provided a good balance between performance and efficiency for our experiments.

\section{Exploring Length-related Issues}\label{sec:sensitivity-seq-len}

\begin{figure}[t]
    \centering
    \includegraphics[width=0.95\columnwidth]{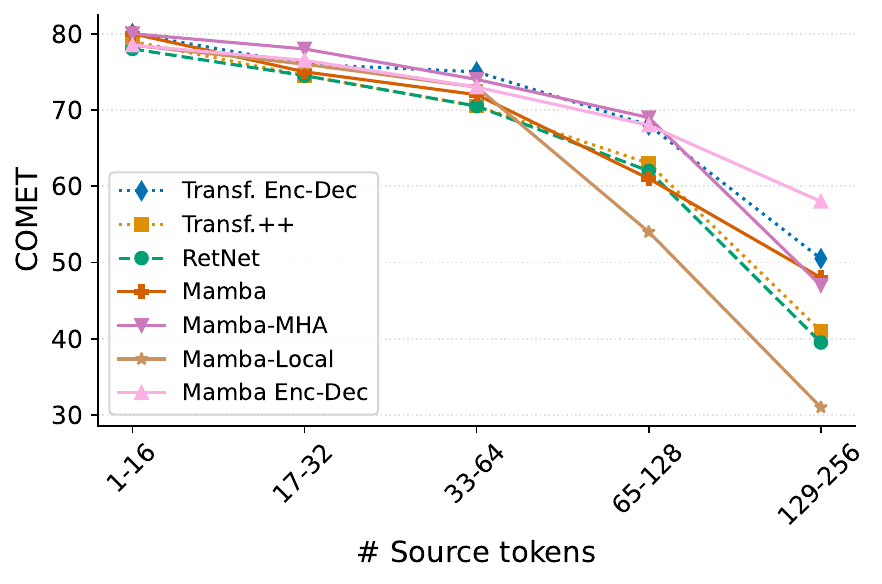}
    \caption{COMET scores per sequence length on WMT14 \textsc{de$\to$en} for trained-from-scratch models.}
    \label{fig:wmt14-comet-lens}
\end{figure}

\subsection{Preliminary Sentence-level Experiments}

Before experimenting with paragraph-level data, we analyze how our trained-from-scratch models perform on different sequence lengths. To this end, we study their sensitivity to input length when trained and tested on WMT14 \textsc{de$\to$en}. The results are shown in Figure~\ref{fig:wmt14-comet-lens}. While all models show a deterioration in performance as sequence length increases, this effect is more pronounced for Transformer++, RetNet, and Mamba-Local, with a significant drop in performance for samples longer than 64 tokens.

\begin{figure*}[t]
    \centering
    \includegraphics[width=\textwidth]{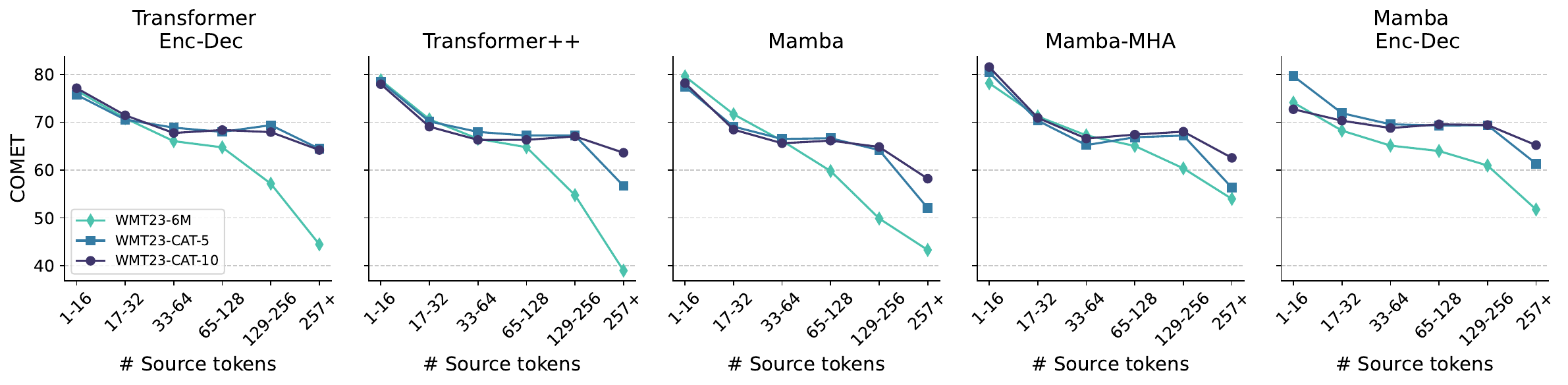}
    \includegraphics[width=0.96\textwidth]{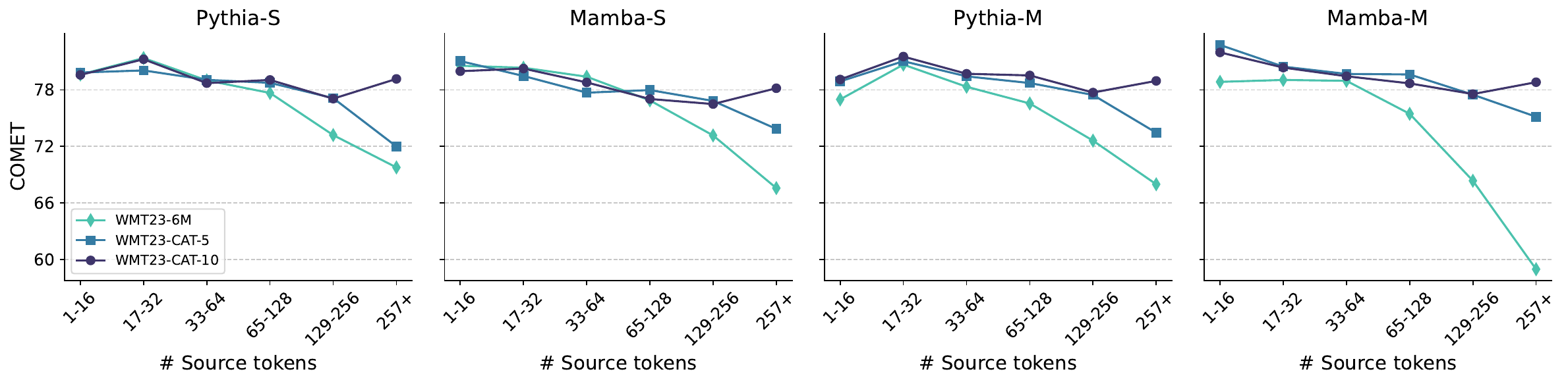}
    \captionof{figure}{Sensitivity to input length, measured by the number of sources tokens, on the WMT23 \textsc{en$\to$de} datset, for models trained from scratch (top) and finetuned from a pretrained checkpoint (bottom).
    }
    \label{fig:ende-ms-extrapolation}
\end{figure*}

\subsection{Sensitivity to Input Length}
\begin{figure*}[t]
    \centering
    \includegraphics[width=0.44\textwidth]{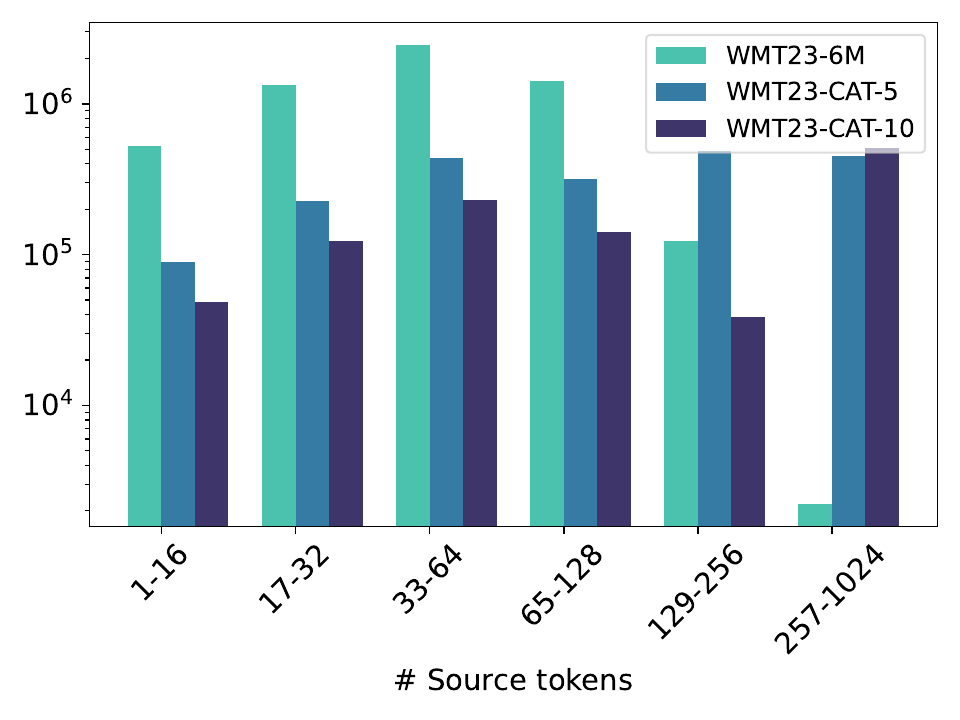}%
    \includegraphics[width=0.44\textwidth]{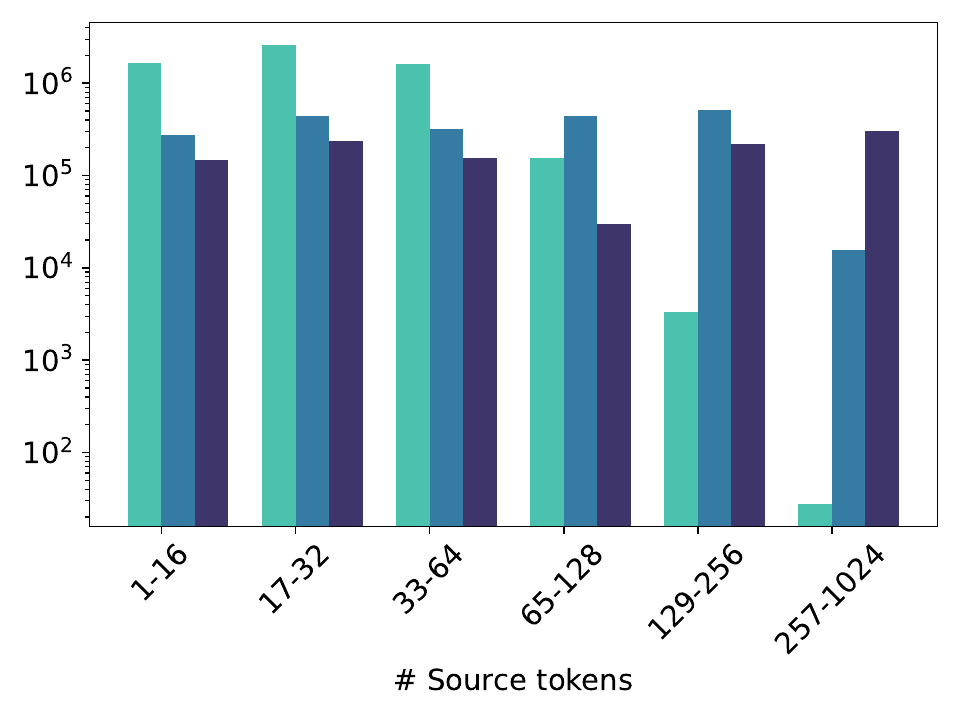}%
    
    \includegraphics[width=0.44\textwidth]{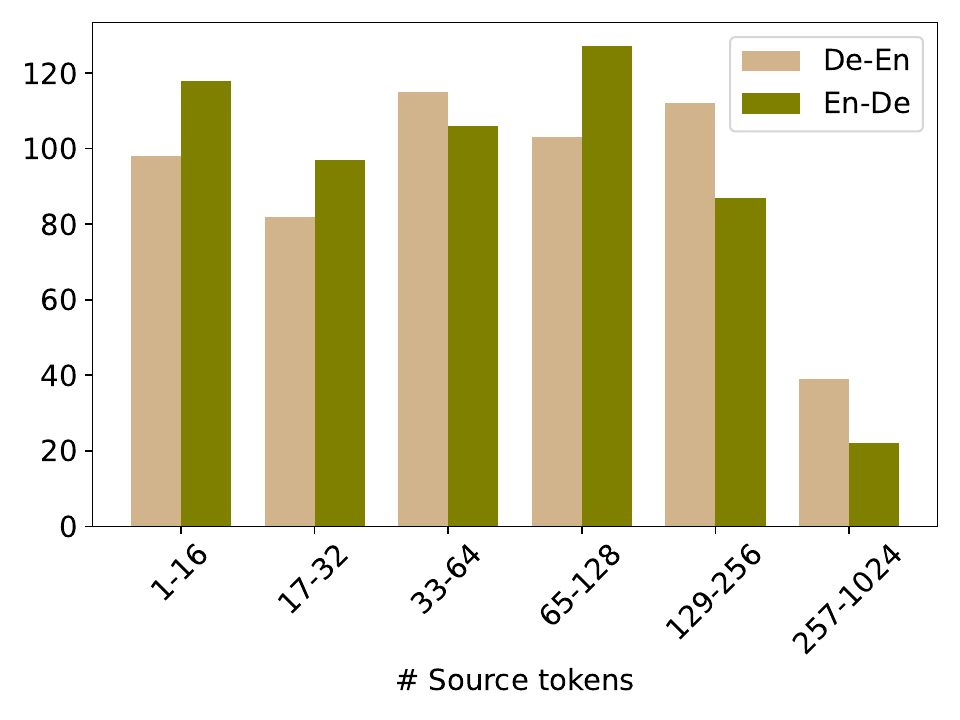}
    \includegraphics[width=0.44\textwidth]{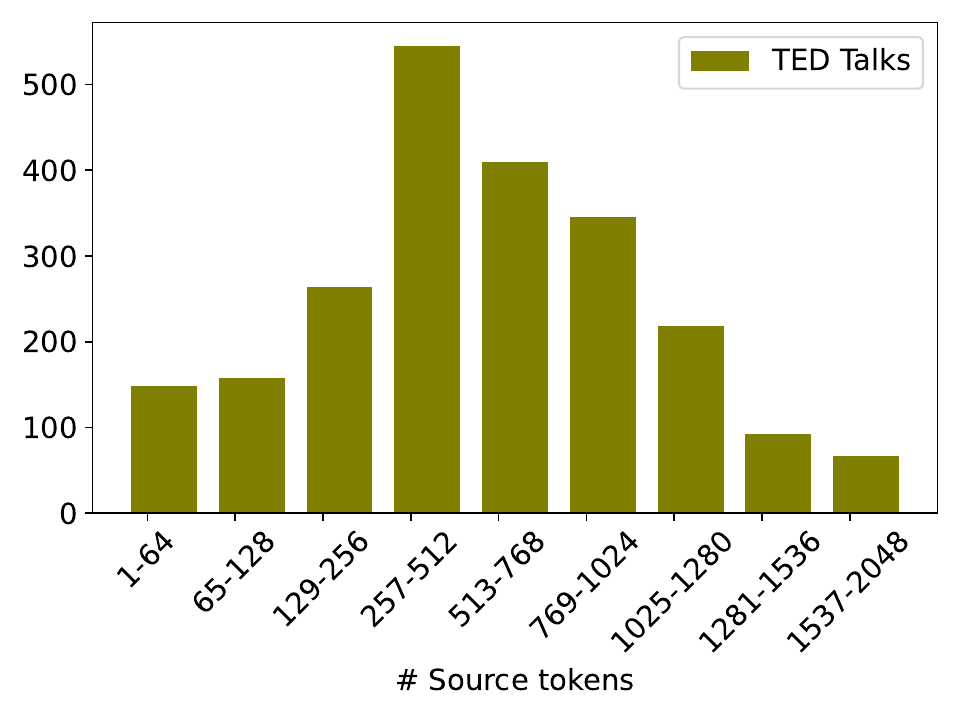}
    \caption{Distribution of source length in 1) the training datasets: WMT23 \textsc{de$\to$en} (top left), WMT23 \textsc{en$\to$de} (top right), and 2) the test datasets: WMT23 \textsc{de$\leftrightarrow$en} (bottom left), our custom TED Talks \textsc{de$\to$en} (bottom right). %
    }
    \label{fig:wmt23-distribution}
\end{figure*}

Following the discussion in \S\ref{extrapolation-issues}, we further investigate the sensitivity of our models to input length using the WMT23 \textsc{en$\rightarrow$de} test set, with results shown in Figure~\ref{fig:ende-ms-extrapolation}. Notably, our takeaways remain broadly the same: concatenating samples in the training data is indeed helpful when handling longer sequences, and models trained on the WMT23-CAT-10 dataset are much better in the longer bin (257+) with minimal translation quality degradation in shorter samples.
However, when considering each of the training datasets' histograms in Figure~\ref{fig:wmt23-distribution}, we can observe that models have been exposed to the longest samples during training, even if in low quantities. 
This implies that the previous experiments do not represent an extrapolation setting, where inference is done on longer sequence lengths than those seen during training. 
We cover extrapolation to longer sequences next.

\subsection{Extrapolation to Longer Sequences}\label{extrapolation}

\begin{figure*}[t]
    \centering
    \includegraphics[width=\textwidth]{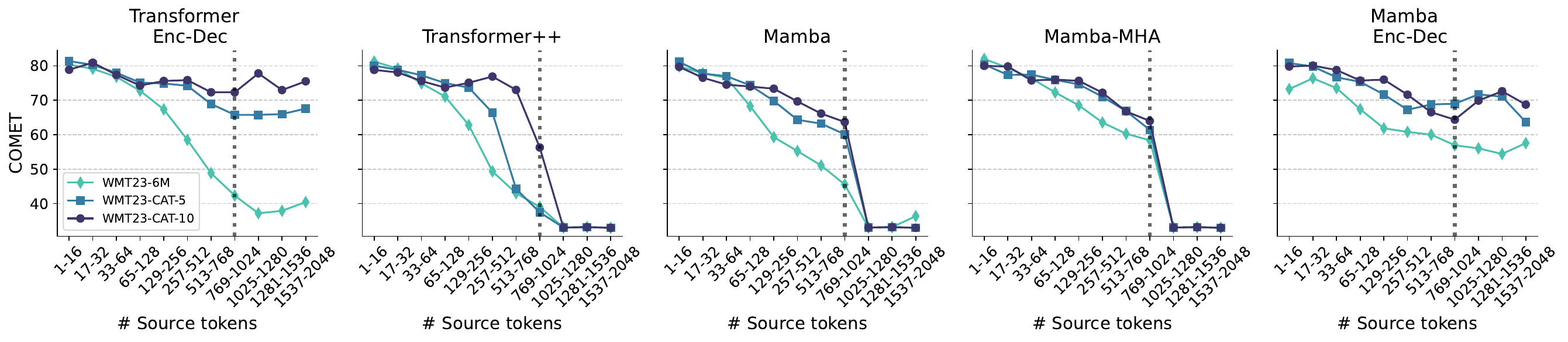}
    \includegraphics[width=\textwidth]{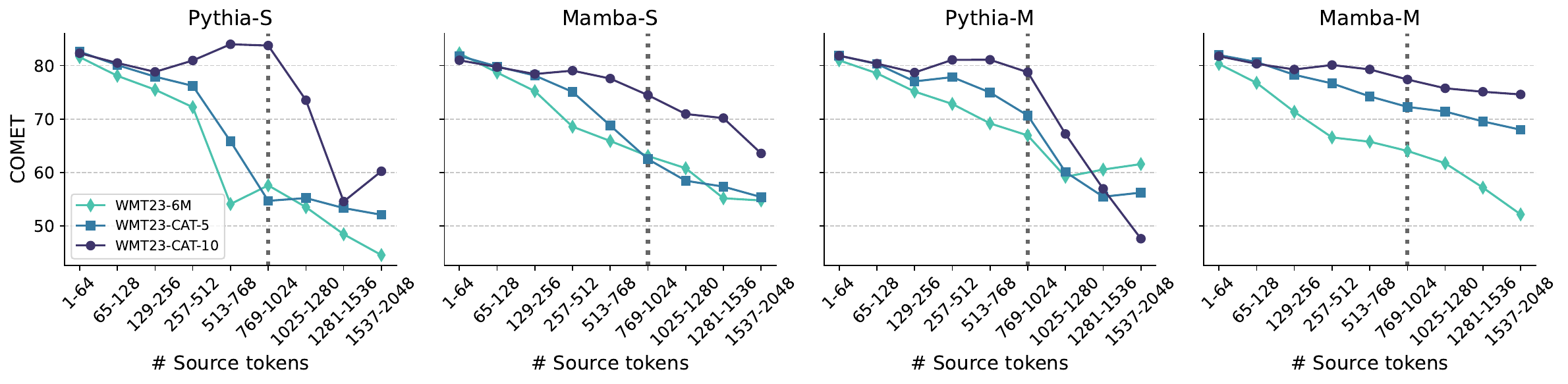}
    \captionof{figure}{Sensitivity to input length, measured by the number of sources tokens, on the Ted Talks \textsc{de$\to$en} dataset, for models trained from scratch (top) and finetuned from a pretrained checkpoint (bottom). 
    The dashed vertical line represents the bin containing the longest sentence in the training set.
    }
    \label{fig:ted-comet-lens}
\end{figure*}

Following the previous discussion, to further explore the impact of sequence length on our models, we create a new test set sampled from TED Talks \textsc{de$\to$en} passages that is much larger (2200 samples) and contains much longer sequences. 
The source length distribution can be seen in Figure~\ref{fig:wmt23-distribution} (bottom right).
After evaluating our models in this dataset, we plot COMET scores per sentence length in Figure~\ref{fig:ted-comet-lens}. Note that the dashed vertical line represents the bin containing the longest sentences the model has been exposed to during training.

\paragraph{Discussion.} 
We observe some interesting behavior: when training from scratch, the translation quality of Transformer++, Mamba, and Mamba-MHA falls sharply when handling 769+ tokens, whereas Mamba Enc-Dec excels even in pure extrapolation settings on the longest inputs. 
With the finetuned models, 
we also see decreasing translation quality over longer sequences, consistent with previous experiments. 
Nonetheless, Mamba models show a more robust trend. In particular, Mamba-M extrapolates well to longer sequences when trained on CAT-10.
For example, for models trained on CAT-10, the best COMET score for inputs longer than 1024 tokens for Pythia-M is $\sim$68, while Mamba-M is able to achieve a score of $\sim$75. The gap increases and reaches almost 20 points as we increase the sequence length.

\section{Full Paragraph-Level Results}\label{appendix:full-paragraph}

For completeness, we report paragraph-level results in terms of BLEU and COMET for all language pairs and models in Table~\ref{tab:paragraph-mt-results}.

\begin{table*}[!htb]
\centering
\begin{tabular}{lccccccc}
\toprule
& & \multicolumn{2}{c}{\sc de$\to$en } & \multicolumn{2}{c}{\sc en$\to$de} \\
\cmidrule(l){3-4} \cmidrule(r){5-6} 
\sc Model &  \sc Training data & \sc BLEU& \sc COMET &  \sc BLEU & \sc COMET  \\
\midrule
\textit{Trained from scratch} \\
\text{Transformer Enc-Dec} & \multirow{5}{*}{WMT23-6M} & 25.4 & 72.4 & 22.4 &  65.2  \\
\text{Transformer++} & & 21.6  & 70.7  & 20.2  &  64.8  \\
\text{Mamba} &  & 19.0 & 70.0 & 15.8  &  63.3 \\
\text{Mamba-MHA} &  & \textbf{23.9} & \textbf{72.7} & \textbf{23.2} & \textbf{67.0} \\
\text{Mamba Enc-Dec} & & 22.7 & 70.7 & 21.5 & 65.3  \\

\cdashlinelr{1-6}

\text{Transformer Enc-Dec} & \multirow{5}{*}{WMT23-CAT-5} & \textbf{30.8} & \textbf{74.6} & \textbf{29.9} & 70.3  \\
\text{Transformer++} & & 28.9 & 73.6 & 28.1 & 69.1 \\
\text{Mamba} &  & 26.1 & 73.3 & 23.8 & 67.5 \\
\text{Mamba-MHA} &  & 29.5 & 74.2 & 23.5 & 68.6 \\
\text{Mamba Enc-Dec} & & 27.3 & 73.8 & 29.1 & \textbf{71.0}  \\

\cdashlinelr{1-6}
\text{Transformer Enc-Dec} & \multirow{5}{*}{WMT23-CAT-10} & 28.3 & 69.6 & 29.3 & \textbf{70.3}\\
\text{Transformer++} & & 29.8 & 72.8 & 29.1 & 68.8 \\
\text{Mamba} &  & 25.9 & 72.3 & 25.5 & 67.8 \\
\text{Mamba-MHA} &  & 27.8 & 74.5 &  25.9 & 69.7 \\
\text{Mamba Enc-Dec} &  & \textbf{31.4} & \textbf{75.6} & \textbf{30.1} & 70.1 \\

\midrule
\textit{Finetuned} \\
\text{Mamba-S} & \multirow{4}{*}{WMT23-6M} & 21.8 & 77.2 & 21.4 & 72.4 \\
\text{Pythia-S} &  & 23.9 & \bf 77.4 & \bf 25.9 & \bf 76.7 \\
\text{Mamba-M} & & 20.7 & 74.6 & 22.5 & 73.4 \\
\text{Pythia-M} & & \bf 26.0 & 76.2 & 25.2 & 75.8 \\

\cdashlinelr{1-6}
\text{Mamba-S} & \multirow{4}{*}{WMT23-CAT-5} & 24.3 & 78.2 & 23.3 & 74.2 \\
\text{Pythia-S} &  & \bf 27.0 & 78.4 & \bf 28.6 & \bf 77.8 \\
\text{Mamba-M} & & 26.4 & \bf 79.6 & 27.5 & 77.5 \\
\text{Pythia-M} & & 25.8 & 78.6 & 27.5 & 77.4 \\

\cdashlinelr{1-6}
\text{Mamba-S} & \multirow{4}{*}{WMT23-CAT-10} & 25.6 & 78.3 & 22.5 & 73.1 \\
\text{Pythia-S} &  & 26.8 & 79.0 & 29.3 & 77.1 \\
\text{Mamba-M} & & 32.5 & \bf 79.5 & 27.5 & 77.3 \\
\text{Pythia-M} &  & \bf 33.4 & 79.4 & \bf 33.9 & \bf 79.0 \\
\bottomrule
\end{tabular}
\caption{Paragraph-level results in terms of BLEU and COMET on the WMT23 \textsc{en$\leftrightarrow$de} test set.}
\label{tab:paragraph-mt-results}
\end{table*}

\section{AI assistants}
We have used Github Copilot\footnote{\url{https://github.com/features/copilot}} during code development, and ChatGPT\footnote{\url{https://chat.openai.com/}} during paper writing for paraphrasing or polishing original contents.

\end{document}